\def\year{2018}\relax
\documentclass[letterpaper]{article} 
\usepackage{aaai18}  
\usepackage{times}  
\usepackage{helvet}  
\usepackage{courier}  
\usepackage{url}  
\usepackage{graphicx}  

\usepackage{booktabs}

\usepackage{subfigure} 
\usepackage{capt-of}
\usepackage{algorithm}
\usepackage{algorithmic}
\usepackage{amsmath}
\usepackage{amsfonts}
\usepackage{blindtext}
\usepackage{hyperref}

\usepackage{comment}

\usepackage{amsthm}

\begin{document}


\title{Learning to update Auto-associative Memory in Recurrent Neural Networks for Improving Sequence Memorization}
\author{Wei Zhang, Bowen Zhou\\
AI Foundations, IBM Research\\
Yorktown Heights, NY\\
{zhangwei,zhou}@us.ibm.com
}
\maketitle

\begin{abstract} 
Learning to remember long sequences remains a challenging task for recurrent neural networks. Register memory and attention mechanisms were both proposed to resolve the issue with either high computational cost to retain memory differentiability, or by discounting the RNN representation learning towards encoding shorter local contexts than encouraging long sequence encoding. Associative memory, which studies the compression of multiple patterns in a fixed size memory, were rarely considered in recent years. Although some recent work tries to introduce associative memory in RNN and mimic the energy decay process in Hopfield nets, it inherits the shortcoming of rule-based memory updates, and the memory capacity is limited. This paper proposes a method to learn the memory update rule jointly with task objective to improve memory capacity for remembering long sequences. Also, we propose an architecture that uses multiple such associative memory for more complex input encoding. We observed some interesting facts when compared to other RNN architectures on some well-studied sequence learning tasks.
\end{abstract}

\section{Introduction}
In recent years, recurrent neural networks (RNN) has been widely used in deep-learning solutions for many real-world tasks. However, RNNs are usually challenged to encode long sequences that popular RNNs choices may not be able to handle easily. For example, Machine reading comprehension, text entailment, and neural machine translation always require long sentences or passages as input;  text generation tasks, such as image captioning and text summarization, always require RNNs to generate many output words one at a time; reinforcement learning tasks, such as dialogue systems, planning and control, frequently uses RNNs as policy networks for action sequence generation. The delayed reward signal mandates RNNs to retain and intelligently compose historical information in memory through time.

However, RNN accumulates input information in a fixed-size vector. Despite the understanding that RNNs such as LSTM \cite{hochreiter1997long} are Turing-complete and could theoretically simulate any function \cite{siegelmann1995computational}, in reality, it is hard for RNN to encode long sequences due to the complexity of the problem. For example, Deep LSTM reader \cite{hermann2015teaching} built for machine reading comprehension task, uses LSTM to simulate how human answer the question: read the passage and the question sequentially word-by-word, with a single LSTM chain. It composes all passage and question words in a single hidden state, and then infer the answer from the state. But the accuracy is relatively low. The effective methods for reading comprehension usually alleviate the stress on LSTM and compare the question representation directly to each passage word representation via attention mechanism. 

There are mainly two ideas that enhance RNNs to deal with long inputs: \textit{attention mechanism} and \textit{external memory}.

\textit{Attention mechanism}, specifically in NLP applications such as neural machine translation\cite{bahdanau2014neural} and machine comprehension \cite{hermann2015teaching}, is external memory built on top of RNNs. For example, in machine reading comprehension, a passage and the related question will be encoded separately with two RNNs, so that each word will have a corresponding hidden state. Then, the each passage word hidden state and the question hidden state (aggregation of all question word hidden states) will be composed to generate an attention score, denoting the similarity of the passage word state (encoding the context around the word) and the question. The attention mechanism is a de facto industry standard that helps to achieve state of the art accuracy on datasets such as Stanford Question Answering Dataset \cite{rajpukar2016}. Although effective, the network for attention mechanism scales up linearly with the input, and it may be computationally inefficient when the network is complex. 

\textit{Register Memory models}, such as Neural Turing Machines \cite{graves2014neural}, introduce a  mechanism to enhance recurrent neural networks by distributing inputs into a differentiable, ``external'' memory block to alleviate state compression. In NTM, a copy of memory state has to be preserved at each time step, and the memory addressing mechanism that controls reading and writing in the memory block may not always lead to well-distributed content, thus reducing the effect of the memory block. Those problems impair NTM families popularity in resolving real-world problems that involve long input sequences.

Some other ways to enhance RNN could be by naively increasing networks layers, inserting intermediate steps between adjacent time steps, adding bi-directional encodings, or increasing hidden unit size. Those extensions, although leading to some improvements, are limited to the power of the structure of the underlying recurrent cell.

We would investigate how attention or memory could remember long sequences more efficiently, that is, when the number of inputs scales up, the RNN does not need to increase its memory as much as the approaches mentioned above do. In fact, the answer has been there for decades.

\subsection*{Associative Memory as Efficient RNN Working Memory}

First proposed by \cite{kohonen1974adaptive} and later on populated by \cite{hopfield1982neural},  Associative Memory models, such as Hopfield nets, studies the problem of storing multiple input patterns in a single network of fully interconnected artificial neurons. The main research question is to increase network capacity, i.e. the maximum number of input patterns that can be stored and almost perfectly recovered, which is controlled by two following aspects: 1) \textit{neuron connection weight matrix} that defines the connection strength between a pair of neurons.  2) \textit{memory update rule} that controls how to modify the states (values) of the neurons collectively, so that the total energy of the network can be minimized (i.e., the system reaches a stationary point), by using the weight matrix.

Hopfield nets are optimized by unsupervised maximum likelihood estimation towards a minimum system energy for storing the collection of input patterns. A settling (learning) process (similar to stochastic gradient descent) is conducted to update the network memory states, while the neuron connection weight matrix is predefined and stay unchanged throughout the course. As a continuous version of associative memory, Deep Boltzmann Machines \cite{smolensky1986information}\cite{salakhutdinov2009deep}, Restricted Boltzmann Machines \cite{nair2010rectified}, Recurrent temporal RBM \cite{sutskever2009recurrent} share the concept of such a settling process for unsupervised feature learning. The objective of the process (i.e. maximizing the probability $P(X)$ of the data $X$) is independent to the objective of the end goal of the network (maximizes probability $P(Y|X)$ fo label $Y$ conditioned on $X$). Learning $P(X)$ has been claimed to help with some tasks such as image classification, though it was not a widely observed fact. Also, Boltzmann Machine has to perform sampling to estimate the partition function for gradient calculation, which inevitably introduces high computational cost. 

Making Hopfield nets or Boltzmann Machines recurrent was investigated before. For example, RTRBM or DyBM \cite{osogami2015learning} merges the energy minimization objective with the Back Propagation Through Time algorithm which is seen as a supervised task-specific fine-tuning, into an integrated learning procedure. For example, to use RBM for image classification, we could first pre-train RBM weights towards minimizing the Associative Memory system energy to learn the ``representation of image'', and later on use the learned weights as a start point to fine-tune networks towards the task objective, e.g., image classification. The pre-training, although maximizing the utility of the training data by discovering ``shape invariant'' representation with BM, is usually off the classification target, and takes much longer time due to sampling for gradient estimation. Our work naturally addresses this issue by retargeting the objective of settling process to the supervised task objective.

It is favorable to learn the associative memory updates with BPTT. An energy annealing phase may be needed, but as suggested by \cite{hinton2010practical}, under-fitting memory to a higher energy and tune the system more with fine-tuning (back propagation) towards the supervised task goal is beneficial. Admittedly, such thinking could diminish the effect of ``unsupervised feature learning''. But it is not guaranteed that an elaborated feature learning could be a success in different problems. In this paper, we create a learn-able memory accumulation rule, which is different from but resembles the energy minimization rule, that is tuned towards the task objective via BPTT. We argue this method can align the idea of associative memory learning and the task fine-tuning while preserving the strength of the associative memory architecture.

Now, it is the question of where such a memory can be stored in an RNN. We would not create an external memory because the additional transformations for memory addressing are necessary between RNNs and memory component. Associative memory does not try to perform explicit addressing though, and it is easier to integrate them into RNNs. 

RNN cells do provide room for such a requirement. Let us think about the weights put on hidden states in LSTM, for example. The gates $i,o,f= \sigma(W_xx+W_hh+b)$ contains $W_x\in \mathbb{R}^{I \times H}$ and $W_h\in \mathbb{R}^{H \times H}$ that are used to balance the contribution between input $x$ and hidden state $h$. $W_x$ may be sufficient for the purpose, and $W_h$ just introduces additional transformation to map $\mathbb{R}^H$ to another vector of the same shape. It is convenient to use $W_h$ for Associative Memory storage, without adding a new term, and call it weight matrix $A$.

There are various ways to create memory $A$. Notably, \cite{hinton1987using} introduces the concept of ``fast weights'', and form $A$ by composing fast (Change with time step) and slow (Standard) weights to deblur images by providing a partial image as input, an effect similar to Hopfield net. Very recently, \cite{ba2016using} creates $A$ by using ``fast weights'' that accumulates the outer-product of the hidden states. $A$ is learned end-to-end with back-propagation with supervised task objective, and the method can be understood as an attention over history. Although \cite{ba2016using} tries to mimic Hopfield nets energy minimization through multi-step memory processing with intermediate RNN steps, there is still no MLE involved. Thus the multi-step process shows little effect. However, the accumulation of input hidden state outer-product is interesting, since it provides a recipe for storing sequence data, which is different from recurrent Boltzmann machines or Hopfield nets. In such setting, the memory connection weight matrix which is defined by the interaction of input hidden states is changing through time, but the memory update rule, which is critical to memory learning, is still rule-based. 

To remember long sequences, Changing memory connection weights to ``higher order‘’, instead of the 2nd order used in Hopfield Nets \cite{krotov2016dense} \cite{demircigil2017model} could also help increase memory capacity, i.e., the connection weights are composed of 3 or more input bits instead of 2 as in Hopfield nets or BM. The same update rule as in Hopfield nets could be applied. The intuition is that higher order weights could distribute input patterns in a larger space than 2nd order weights so that collision of patterns could be alleviated. However, such an expansion to higher order connection matrix is not an elegant solution since the size of associative memory weight will grow exponentially with the increase of the order of weights.

Instead of increasing the order of neuron connection weights, neuron update rule may also lead to the change in the memory capacity. There has been abundant research on manually defined update rules for Hopfield Nets. Boltzmann Machines.
But learning the memory updating end-to-end with RNN. Thus the associative memory update process and task-oriented fine-tuning (BPTT) are aligned. Compared to the latest Fast Weights work \cite{ba2016using}, we make a fundamental and intuitive change that effectively increased the memory capacity. Meanwhile, the difference in network design is embarrassingly small.

To increase the memory capacity non-linearly, we also use several such associative memory blocks and use an addressing mechanism to control which block the inputs are routed dynamically. Linearly increasing such cells could lead to more interesting non-linear storage patterns for inputs of different ``types''. The routing mechanism generates an attention over associative memory blocks.

Combining contributions mentioned in previous two paragraphs, we created an RNN architecture that 1) uses associative memory as working memory that reuses excess weights in RNN, and uses learned memory update rule to increase memory capacity; 2) scales up to an array of associative memory blocks to enhance input representation. Next session, we are going to discuss in detail the two contributions in section \ref{memupdate}
and \ref{weinet_arc}

\section{Learning Autoassociative Memory Updates}
\label{memupdate}

Before diving into the architecture, we describe the learned associative memory update in RNN that is critical to the success of our model. Closely related to our work, the new Fast Weights work \cite{ba2016using} introduced auto-associative memory into RNN by accumulating the outer-product of RNN hidden state to form the memory. The recurrent update is $\mathbf{h}_{t+1} = tanh( W\mathbf{x} + A_t\mathbf{h}_t + \mathbf{b})$, and $A$ is the accumulation of the outer-product as $A_t = \lambda A_{t-1} + \eta \mathbf{h}_t \otimes \mathbf{h}_t$, where $\otimes$ denotes the outer-product, $\lambda, \eta$ are scalars, $tanh$ for hyperbolic tangent function, $\mathbf{h},\mathbf{b}\in \mathcal{R}^{H}$, $A \in \mathcal{R}^{H\times H}$, $W \in \mathcal{R}^{I\times H}$, $H,I$ are hidden unit size and input dimension. Although driven by the idea of associative memory, the rationale of Fast Weight was explained by attention mechanism, where $A_t\mathbf{h}_t$ is the attentive sum of $\mathbf{h}_t$ in history. The effect of Fast Weights is explained by correspondence to Hopfield net and the neurobiological intuition.

Krotov and Hopfield \shortcite{krotov2016dense} explain that the capacity of Hopfield net is bounded by the speed of energy decay related to second-order neuron connection weight. This could explain why RNN is inferior: The hidden state is first-order, which means no cross bit connection (such as term $h_t^i*h_t^j$ in Hopfield net) is defined in $\mathbf{h}_t$. The intuition is that the higher the order of connection is, the easier and the more the inputs can be stored. For example, when input vectors are orthogonal, higher order memory is easier to learn to distribute inputs to different locations in memory to resolve input conflict than first-order vector memory, so that capacity of the memory can increase. Thus, \cite{ba2016using} which uses a second-order auto-associative memory could have better chance storing and recalling more input contents. However, \cite{krotov2016dense} also experimented with higher order ($>$3) memory connection weights, and showed non-significant improvement over second-order one. The observation indicates that connection matrix could easily hit a storage limit with Hopfield-net-like memory update rule, and a better update mechanism is critical to enable a better memory storage.

\cite{ba2016using} uses a fixed update rule of two scaler hyperparameters to control how fast the accumulation decays ($\lambda$), as well as how much the new input is discounted ($\eta$) to the accumulated $A$. Although efficient for remembering short inputs, those scalar hyperparameters limit the speed of energy decay in terms of Krotov and Hopfield \shortcite{krotov2016dense}. Specifically, different bits in $A$ has the same decay rate that makes storing ``disentangled'' input representation harder; and the decay rate is fixed in advance, instead of being learned automatically, which limits the efficient exploration of memory update speed for different bits to generate dis-entangled memory storage. 

In our work, we simply change the update rule, specifically, the two hyperparameters $\lambda$ and $\eta$ in \cite{ba2016using}, into learnable weight matrices $W_A, W_h\in \mathcal{R}^{H\times H}$. Then, the update rule becomes 
\begin{align}
A_t = W_A \odot A_{t-1} + W_h \odot \mathbf{h}_t \otimes \mathbf{h}_t
\end{align}
Where $\odot$ is the Hadamard product, i.e., bit-wise multiplication. Again, this form, although simple, fundamentally changed how the $A_t$, the associative memory, accumulates. We hypothesize that the weight $W_A$ and $W_h$ could change the ``direction'' of each stored input towards being orthogonal to each other, and intelligently distributing inputs in $A_t$ to increase memory capacity. From attention mechanism perspective, we could regard this as a ``parameterized attention'', But using such an angle is very hard to explain the significance of (1) compared to Fast Weights.

Different from using static scalars $\lambda$ or $\eta$, $W_A$ and $W_h$ is adjusted from iteration to iteration through learning, in a way identical to other learnable parameters in the recurrent cell. Obviously, the addition of $W_A$ and $W_h$ introduces some overhead during training correspondingly. But during network testing, the difference is negligible. 

The change from Fast Weight \cite{ba2016using} to eq. 1 is embarrassingly simple and effective. We will start introducing the overall architecture, which we call WeiNet (Learned Connection \textbf{Wei}ght \textbf{Net}works), where our learned associative memory resides.

\section{WeiNet: RNN with an Array of Auto-associative Memory}
\label{weinet_arc}

\begin{figure}[t]
\vskip 0.2in
\begin{center}
\centerline{\includegraphics[width=\columnwidth]{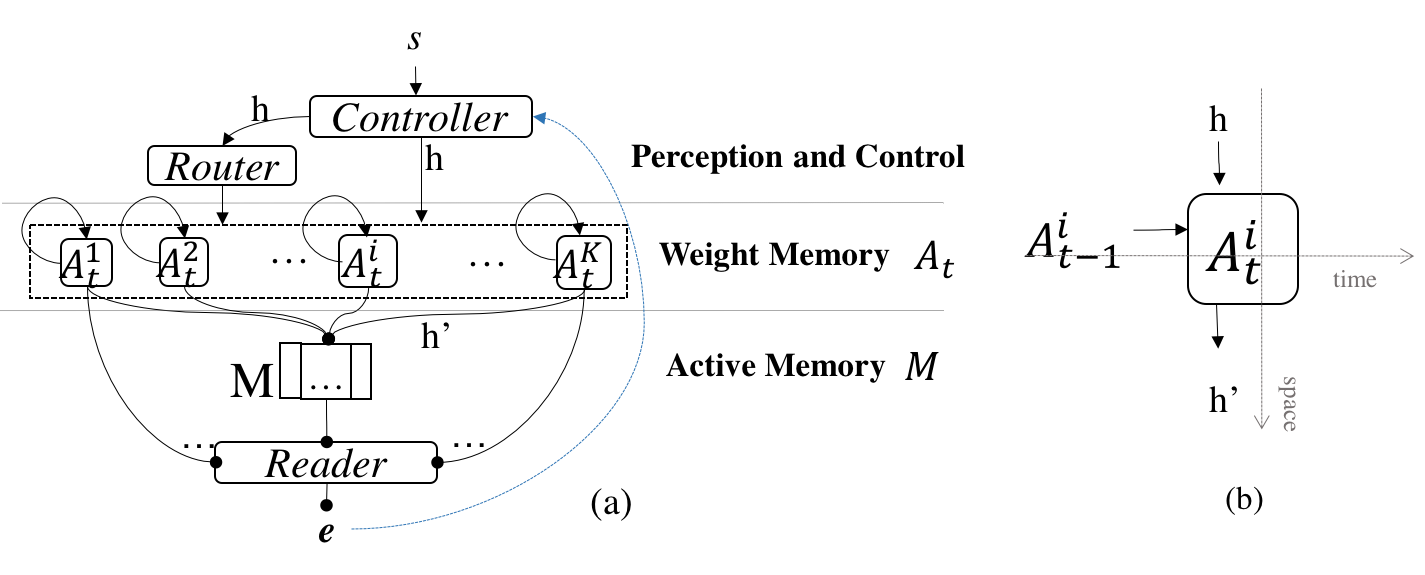}}
\caption{The architecture of WeiNet (a), and the spatialtemporal illustration of Associative Memory (b)}
\label{fig:arc}
\end{center}
\vskip -0.2in
\end{figure} 

WeiNet consists of the following major components shown in Figure ~\ref{fig:arc}(a): 1) \textbf{Input Encoding}: A \textit{Controller} $C$ converts input signal $\mathbf{s}$ into an internal representation $\mathbf{h}$; \textit{Router} $R$ takes $\mathbf{h}$ and soft-select which associative memory to process $\mathbf{h}$. 2) \textbf{Array of Auto-associative Memory (AM) }: An array of $K$ auto-associative memory  $\{\mathbf{A}\}_K$. The self-loop on each $\mathbf{A}^k$ denotes a self-update process as in eq. 1. We use $2D$ matrix for $A$ for this work. 3) \textbf{Active memory $\mathbf{M}$}: Each column is an active memory that trivially stores outputs from previous layer. 4) \textbf{Memory Reader $D$}: Memory reader generates the output of the whole Network, which is read from both auto-associative memory and active memory.

Figure ~\ref{fig:arc}(b) shows how an Auto-associative Memory (AM) $\{\mathbf{A}\}$ impacts $\mathbf{h}$ in both time and space dimension: temporally, AM cell $\mathbf{A}^k$ will be updated to a new state; spatially $\mathbf{h}$ goes through the newly updated $\mathbf{A}^k$ for transformation. The whole networks, although looks intimidating, has simple components as described below:

\noindent\textbf{Input encoding:} 
The \textit{Controller} $C$ is parameterized as a recurrent function $f_C$ that takes the input $\textbf{s}_t$, previous hidden state $\mathbf{h_{t-1}}$ and Memory Reader $D$'s output $\mathbf{e}_{t-1} \in \mathcal{R}^{H}$ to generate $\mathbf{h}_t \in \mathbb{R}^H$, where $H$ is the hidden unit size, with:
\begin{align}
\mathbf{h}_t = tanh( W [\mathbf{s}_t; \mathbf{e}_{t-1}; \mathbf{h}_{t-1}])
\end{align}
where $W \in \mathcal{R}^\{H\times I+2H\}$. 
Introducing $\mathbf{e}_{t-1}$ could make $\mathbf{h}_t$ aware of the output from previous reader module. However, $\mathbf{h}_{t-1}$ can also be understood as a highway connection \cite{zilly2016recurrent} that shortcuts the $\mathbf{e}_{t-1}$ generated from the full transformation pipeline in the RNN cell. 
We found that using LSTM cells for eq. 2 updates leads to significant learning slow-down, which may be due to the difficulty of learning gates in LSTM. 
To mitigate gradient vanishing problem, we applied layer normalization \cite{ba2016layer} right after generating $\mathbf{e}_{t-1}$ to regularize the output before passing to next time step.

\noindent\textbf{Routing in auto-associative memory array}: The \textit{Router} R is a recurrent function:
\begin{align}
\mathbf{a}_t &= softmax ( g(\mathbf{A}_t, \mathbf{h}_t) + \mathbf{w} \odot \mathbf{a}_{t-1} )\\
g(\mathbf{A}_t,\mathbf{h}_t) &= [\mathbf{h}_t'\mathbf{A}^1_t\mathbf{h}_t, ..., \mathbf{h}_t'\mathbf{A}^K_t\mathbf{h}_t]
\end{align}

The idea behind eq. 4 is to learn to interpolate a new attention over $\{\mathbf{A}\}$ with $\mathbf{h}_t$ (first term in softmax), and the old attention (second term in softmax). $\mathbf{w}$ is used to balance the contribution between the two terms. $\mathbf{h}^T\mathbf{A}^k_t\mathbf{h}_t$ in eq. 5 is a scalar, and $g(\mathbf{A}_t, \mathbf{h}_t)$ is a vector of length $K$. 

\noindent\textbf{Auto-associative Memory array update:} eq. 1 provides a simpler version on how we update AM. To make update more efficient, we introduce a cross-talk term between $\mathbf{A}$ and $\mathbf{h}$:
\begin{equation}
\begin{split}
\mathbf{A}_{t}^k = & f_U(W_A \odot \mathbf{A}_{t-1}^k + W_h \odot (\mathbf{h}_t \otimes \mathbf{h}_t)\\
&+ W_{AH} \odot \mathbf{A}_{t-1}^k \odot (\mathbf{h}_t \otimes \mathbf{h}_t))
\end{split}
\end{equation}

$W_A, W_H, W_{AH} \in \mathbb{R}^{H\times H}$. The third term directly models the Hadamard product between $\mathbf{A}_{t-1}$ and $\mathbf{h}_t$ outer product. Such design is driven by effectiveness of two types of attention mechanism: 1) machine-translation-style \cite{bahdanau2014neural} attention by encoder-state and decoder-state weighted addition, and 2) reading-comprehension-style \cite{yu2016end} attention score by state-state dot product. Choosing addition or dot product may depend on whether two hidden states to be compared lies in the same space. In Eq. 5, the third term provides a chance to discover relationships that first two terms do not provide, so that the capacity of associative memory $A$ may be further enhanced. Also notice that in each term the weight $W$ performs only bitwise multiplication instead of matrix production, to make the learning faster, while adequate to serve the purpose of input accumulation. 

The function $f_U$ in this work is simplified to be identity function to increase network training efficiency, which enabled faster convergence than using hyperbolic tangent. We also tried removing the third term to simplify the form and observed slightly worse accuracy across tasks, which confirmed the effectiveness of the formation.

\textbf{Writing to and read from memory:}
The active memory $\mathbf{M}$ simply stores hidden states $\mathbf{h}'$ denoted in \ref{fig:arc}(b). However, we directly use the content in $\mathbf{M}$ and generate a retrieved content 
\begin{align}
\mathbf{m}_t =\mathbf{h}_t \sum_{k=1}^K \mathbf{a}_t^k \mathbf{A}_t^k
\end{align}
When $\mathbf{a}_t$ is a probability distribution over array of $K$ AM, and a single AM $\mathbf{A}_t^k$ will be ``soft-selected'' to generate $\mathbf{m}_t$ when $\mathbf{a}_t$ is a sharp distribution. The dot product between $\mathbf{h}_t$ and the attention sum of AM demonstrates a retrieval process, where $\mathbf{h}_t$ is the key, and the attention sum of $\{\mathbf{A}\}$ is the memory. The attended sum of $\{\mathbf{A}\}$ enables non-linear memory composition not found in a single memory, which is similar to register memory's \cite{graves2014neural} memory slot addressing mechanism. Such mechanism is effective even when addressing cannot be learned\cite{zhang2015structured}.  Eq. 6 can also be regarded as an attention over attention mechanism, where the first one is enforced by router over $\{\mathbf{A}\}$, and second attention is from $\mathbf{h}$ over a series of $\mathbf{A}^k_t$.

Trivially, The memory read module, $D$, is reading from both auto-associative memory and active memory as 
\begin{align}
\mathbf{e}_t = tanh( W [\mathbf{e}_{t-1}; \mathbf{Ac}_t; \mathbf{Ar}_t; \mathbf{m}_t; \mathbf{h}_t])
\end{align}
where $W \in \mathbf{R}^{H\times 5H}$. $\mathbf{Ac}_t, \mathbf{Ar}_t$ denotes column-wise and row-wise mean of the weighted average of $\{\mathbf{A}\}$ using $W_A$ in eq. 5. Adding those two statistics is useful when the hidden state $\mathbf{h}_t$ should be ignored due to its non-significance to the task objective. We also introduce $\mathbf{h}_t$ as highway connection to provide a shortcut to skip all memory components.

\section{Key difference of WeiNet and other associative memory forms}
Hopfield nets and WeiNet has the radically different way of handling inputs. Take image classification task as an example, Hopfield net stores an image in one shot on the net to encode relation among pixels. And multiple images can be remembered by learning rule-based neuron connections and adjusting the memory of the whole dataset. However, WeiNet has to use RNN to handle the same task, where each time step may handle partial image, and compose them together to get a representation of the full image. The whole dataset can be remembered by adjusting only the network weights during training. The RNN recurrence provides a chance to handle complex input patterns, such as sequential data, which Hopfield nets are difficult at handling.

Boltzmann Machines also has recurrent variants such as Temporal Restricted Boltzmann Machines \cite{sutskever2009recurrent} and Dynamic Boltzmann Machines \cite{osogami2015learning}. TRBM models the interaction between visible units (input) and hidden units through a mechanism that is similar to feed-forward neural nets, and the neurons are not fully connected. WeiNet, as well as \cite{ba2016using}, not only models the interactions between different $\mathbf{h}$ at various time steps, but also the interactions of bits within $\mathbf{h}$ through outer-product like Hopfield nets do. Dynamic Boltzmann Machines uses the full neuron connection, and similar to our networks in this sense. However, the connection weights are STDP-driven and are modeled to follow a parameterized exponential decay process. WeiNet's update rule is simply matrices of scalar weights. \cite{ba2016using} uses fixed scalers that are even simpler, but less flexible than WeiNet's counterpart.

\section{Experiments}

\subsection{Associative Recall task}

\begin{figure}[!htb]
\minipage{0.4\textwidth}
\centering 
\includegraphics[width=60mm]{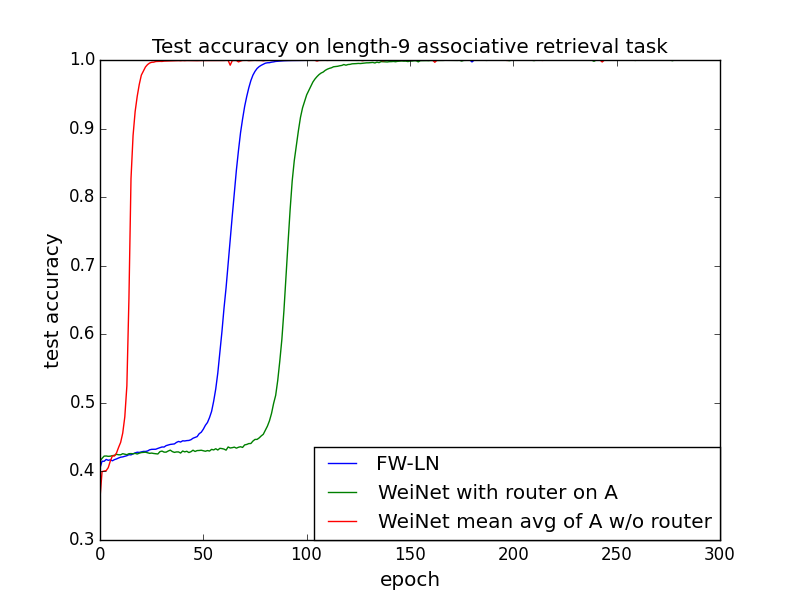}
\label{fig:len_30}
\endminipage

\minipage{0.4\textwidth}
\centering 
\includegraphics[width=60mm]{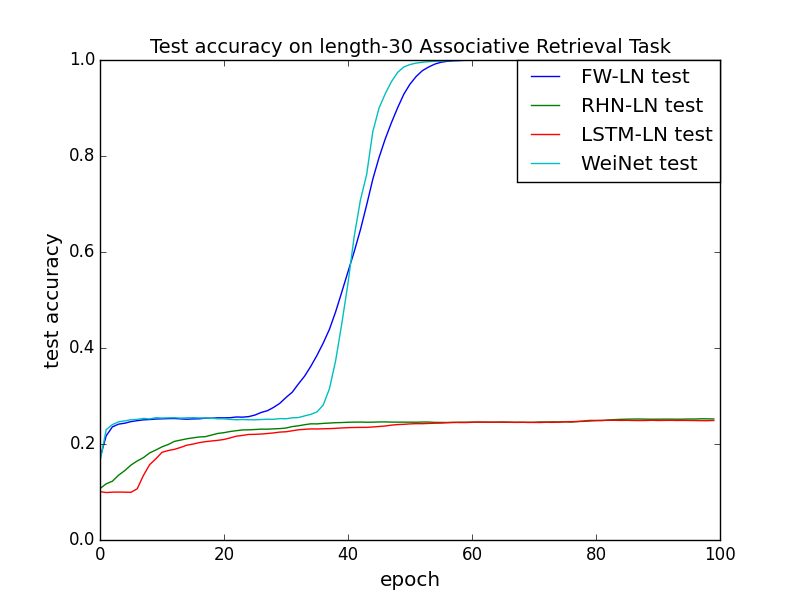}
\label{fig:len_30}
\endminipage\hfill

\minipage{0.4\textwidth}
\centering 
\includegraphics[width=60mm]{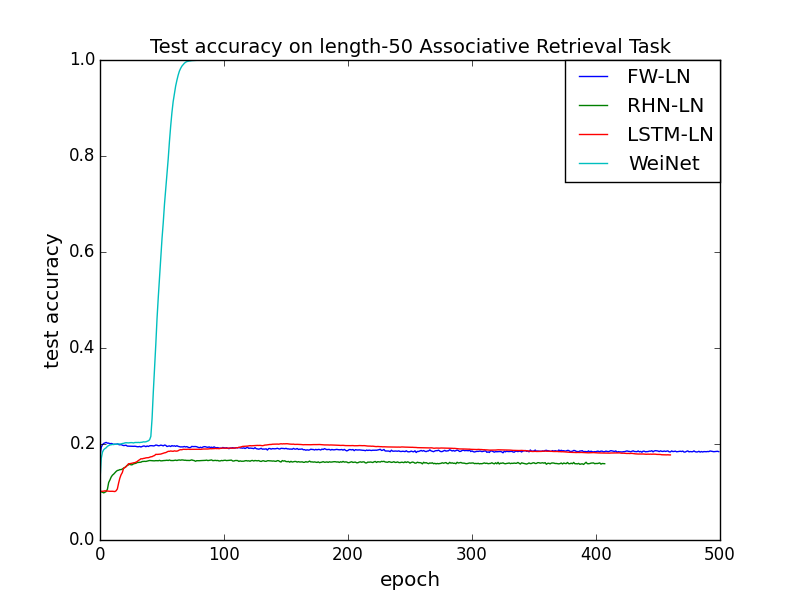}
\endminipage\hfill

\caption{Comparing WeiNet to Fast Weight with Layer Normalization (FW-LN), Recurrent Highway Networks with Layer Normalization (RHN-LN), LSTM with Layer Normalization (LSTM-LN) on Associative Recall Task, using different input lengths.}
\label{fig:ar_len_9_50}
\vskip -0.2in
\end{figure} 

\begin{figure}[!htb]
\minipage{0.4\textwidth}
\centering
\includegraphics[width=60mm]{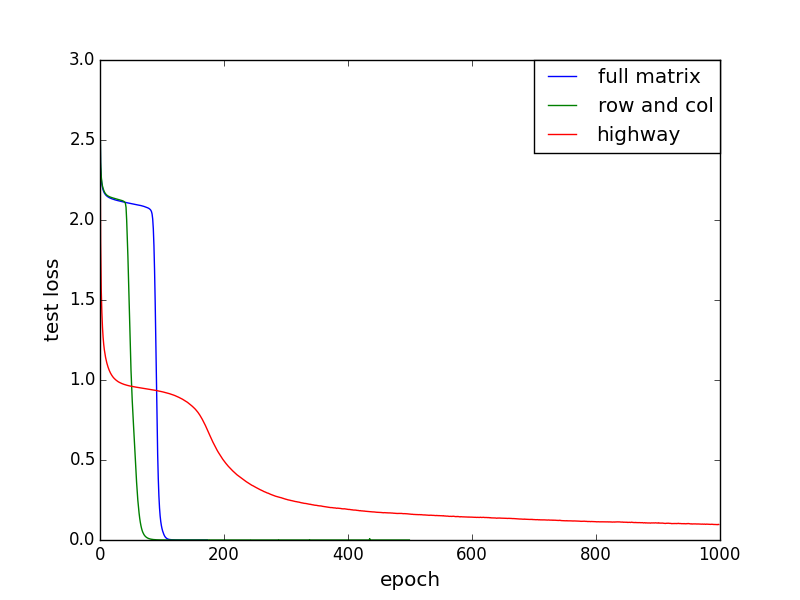}
\caption{Comparison of Different strategies for updating weight memory $\mathbf{A}$ for length-50 associative recall. Row-column vector outer product is easiest for training.}
\label{fig:choice_of_A}
\endminipage
\vskip -0.2in
\end{figure} 
Associative Recall task requires recurrent networks to remember character sequence, and retrieve specific character spatially related to a query character. For example, input sequence to an RNN looks like ``c9k8j3f1??k'', where the sequence to remember is ``c9k8j3f1'', and ``k'' is the key to search in the previous sequence. ``??'' is special character separating the sequence with query ``k''. A neural network is asked to encode the whole concatenated sequence from left to right one character at a time, and then use the last hidden state from RNN after encoding the key ``k'' to predict the retrieved answer ``8''.

To successfully handle the task, recurrent neural networks should be capable of remembering not only input sequence but also the interactions between characters. We would like to test models on a different length of sequences, which may provide insights in model's memory capacity. We use three lengths settings: 9, 30, 50 input characters. For each setting, we create 100K training, 10K validation, and 10K test examples.

We compare WeiNet to LSTM, Recurrent Highway Networks (RHN) \cite{zilly2016recurrent}, Fast Weights with layer normalization \cite{ba2016using} (FW-LN) on all settings. Each model will use 50 hidden units and one recurrent layer. We use ADAM optimizer \cite{kingma2014adam} with learning rate 1e-4, mini-batch size is 128. FW-LN follows setting in \cite{ba2016using}. RHN use coupled gates; LSTM use general architecture without peephole connections, WeiNet uses one fast weight matrix without router networks (Eq. 3 and Eq. 4) to be comparable to the other models. This setting corresponds to Eq. 6 using a single $A$ instead of a weighted sum of $\{\mathbf{A}\}$, and eq. 7 uses the row and column mean from a single memory $A$. This simplified version degrades WeiNet to a structure similar to \cite{ba2016using} with the difference on Eq. 5 for auto-associative memory updates.

Table ~\ref{table:len_9_50} and Figure ~\ref{fig:ar_len_9_50} shows the test accuracy on length-9, length-30 and length-50 sequences. WeiNet achieved 100\% accuracy on all three tasks. Fast Weights with Layer Normalization (FW-LN) fails at the length-50 sequence. We tried using 100 unit for Fast Weight, but the model fails to converge on length 50 either. We further modified the Fast Weights model directly to use the learned updates in Eq. 1 and Eq. 5 respectively, and the model converged and achieved nearly 100\% test accuracy. We further removed parameterized update in Eq. 5 in WeiNet and use a rule-based update as in \cite{ba2016using}, and observed difficulty in learning, which confirms the advantage of our learned auto-associative memory update. 

Fig. ~\ref{fig:ar_len_9_50} (top) also shows two critical settings for WeiNet, with and without \textit{Router} R. Router is not necessary for the task. When using Router, we apply two Associative Memory blocks in WeiNet. Curve shows adding more memory linearly does not help but complicate learning for this task.

Fig. ~\ref{fig:choice_of_A} shows evaluation on different choices of weights for updating $\mathbf{A}$ in eq. 6. We compare 3 approaches: 1) Using $W_A \in \mathbf{R}^{H \times H}$ denoted as ``full matrix''; 2) Using two vectors $\mathbf{w}_c \in \mathbf{R}^{H}, \mathbf{w}_r \in \mathbf{R}^{H}$ to generate an outer product $\mathbf{w}_c \otimes \mathbf{w}_r $ for $W_A$ denoted as ``row and col'' and 3) Using recurrent highway connections for updating $\mathbf{A}$ by introducing coupled gates $g=sigmoid(W_A\odot A_{t-1} + W_h \odot \mathbf{h}_t \otimes \mathbf{h}_t)$. Then, $A_t = g \odot A_{t-1} + (1-g) \odot \mathbf{h}_t \otimes \mathbf{h}_t$. The curves are on length-50 recall task, and we can see that using approach 2) not only significantly reduce the parameter size from $O(H^2)$ to $O(H)$ but also achieves faster convergence thanks to the more efficient training. The observation complies with the intuition that $\mathbf{A}$  has $O(H)$ degrees of freedom, and a parameter of the same degree-of-freedom may suffice. The low accuracy of RHN and LSTM shows that the difficulty of learning sigmoid gates slows down significantly the memory storage, which confirms that the memorization of WeiNet is a combined effect of both input representation learning and memory update learning. Without sigmoid gates and highway connection, WeiNet uses a matrix instead of gates and fight against vanishing gradient by applying gradient clipping of [-5, 5] and layer normalization right after generating $\mathbf{e}_t$, which is proved to be effective in the task. WeiNet runs slower than FW-LN, due to the calculation of gradients for the learning parameters in Eq. 6. However, WeiNet converges with fewer epochs than FW-LN, which reduces the total running time. 

\begin{table*}[t]
\centering
\small
\setlength\tabcolsep{5.5pt}
\begin{tabular}{lcccccccccc}
\toprule
                  & \multicolumn{2}{c}{\bf Length 9 } & \multicolumn{2}{c}{\bf Length 30} & \multicolumn{2}{c}{\bf Length 50} \\
                  & \bf Epochs            & \bf Accuracy            & \bf Epochs            & \bf Accuracy            & \bf Epochs               & \bf Accuracy        \\
\midrule
LSTM & $<$200 & 100\% & 5000 & 25.6 & 5000 & 20.5\% \\
RHN \cite{zilly2016recurrent} & $<$150 & 100\%  & 5000 & 25.7 & 5000 & 18.3\% \\
FW-LN \cite{ba2016using} & $<$30 &100\% &$<$50 & 100\% & 5000 & 20.8\% \\
\midrule
WeiNet         & $<$30 &100\% &$<$35 & 100\% & $<$50  &  100\%\\
\bottomrule
\end{tabular}
\normalsize
\caption{Comparison of models on epochs required to converge and convergence test accuracy}
\label{table:len_9_50}
\end{table*}

\textbf{Explanation on WeiNet's power of learning associative recall} 

As we discussed before, WeiNet does not store a training input in one shot as Hopfield net does, and the way how patterns are encoded is different. For example, if we would learn the associative recall task with Hopfield net, we would let hopfield net to encode each character in a neuron, and the association between neurons could directly encode the relationships between characters. However, WeiNet could also learn relationships in following manner: Let's unroll the simplified WeiNet sequence through time, and assume $A_0=0$. It's not hard to see that
\begin{align}
A_T^{ij} = W_h^{ij} \odot \sum_{t=0}^T (W_A^{ij})^{t-1} \odot (\mathbf{h}_t^{ij} \mathbf{h}_t^{ij} )
\end{align}
We have an assumption that every $A_T^{ij}$ which is the scalar memory at i-th column and j-th row in $A_T$, could encode any two temporally adjacent bits $(\mathbf{h}_s\otimes \mathbf{h}_s)_{ij}$ and $(\mathbf{h}_t\otimes \mathbf{h}_t)_{ij}$ being orthogonal to other adjacent pairs for close time steps s and t. It basically says, using associative recall task example, that any bit $A_T^{ij}$ could learn to store any pair of adjacent characters, because each bit $A_T^{ij}$ is actually a T-th order polynomial function of 
$A_T^{ij}= \mathbf{h}_0^i\mathbf{h}_0^j V^t + \mathbf{h}_1^{i}\mathbf{h}_1^{j} V^{t-1} + ... + \mathbf{h}_{t-1}^{i}\mathbf{h}_{t-1}^{j} V + \mathbf{h}_t^i \mathbf{h}_t^j$ where both $\mathbf{h}$  and $V=W_h^{ij}W_A^{ij}$ are variables. 
In \cite{ba2016using}, only the $\mathbf{h}$ 
are variables, and every bit combination $\mathbf{h}_t^i\mathbf{h}_t^j$ 
has to be mapped to the same space defined by the same set of basis $(V^t, V^{t-1}, ... V, 1)$
Where $V$ is a scalar, thus the bits are more easily entangled and made hard to be stored discretely in auto-associative memory $A$. In WeiNet, there will be different $V$ for various $A_T^{ij}$. Thus the basis is different, especially when the network is randomly initialized. 

In eq. 6, the $W_A$ is a bit-wise scaler of the associative memory $A$. element $\mathbf{A}(i,j)$ of $\mathbf{A}$ is only a function of values at corresponding positions in $\mathbf{A}$ or $\mathbf{h}$, namely $A_t^k(i,j)= f(A_{1}^k(i,j),...,A_{t-1}^k(i,j), h_1^i, ..., h_t^i, h_1^j, ..., h_t^j )$, and no other $A_{\neq{i},\neq{j}}$ is relevant to the making of $\mathbf{A}(i,j)$. We convert the Hadamard product of the first and second term in Eq. 6 into dot product, using the same weight matrices to learn. This enables that each memory $\mathbf{A}(i,j)$ to receive information from all other memory bits. We observed that, when handling length-9 associative recall task, such a change leads to extremely fast convergence of WeiNet. However, when we try on length 30 or length 50, we were not able to see convergence anymore which might be due to over-fitting by introducing cross-bit contribution in the same time step.

\begin{figure}[!htb]
\begin{center}
\includegraphics[scale=0.5]{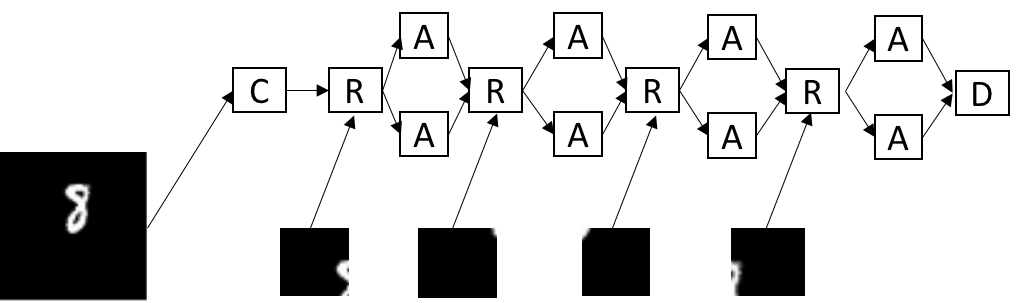}
\caption{Demonstrating how input sequence is fed into WeiNet. C for controller, R for router, D for reader, and A for weight memory. For fixated attention, each small glimpse is $1/4$ of an input image with no pixel overlap. For dynamic attention, each glimpse is generated from glimpse networks and pixel overlap is allowed.}
\label{fig:rva}
\end{center}
\vskip -0.2in
\end{figure}

\begin{table}
\centering
\caption{ Comparison of test set error rate on MNIST classification task. This is result for Setting 1: All models have same hidden size 200, 1 recurrent layer, FW-LN uses 1 single inner recurrent layer. WeiNet uses 1 fast weight memory cell}
\label{table:va_da}
\begin{tabular}{c      c}
\textbf{Model} & \textbf{Test error}\\
\hline
LSTM &0.79 \%  \\
\hline
RHN &0.69\%  \\
\hline
FW-LN & 0.65\% \\
\hline
WeiNet   & 0.49\%  \\
\end{tabular}
\end{table}

\begin{table}
\centering
\caption{ Comparison of test set error rate models on MNIST classification task. This is Setting 2: use glimpses in network inner loop for both FW-LN and WeiNet. Dynamic attention is generated from glimpse networks; fixated attention is using 4 7$\times$7 sub-images}
\label{table:va_fa_vs_da}
\begin{tabular}{c        c}
\textbf{Model} & \textbf{Test Error}\\
\hline
FW-LN (Fixated attention) \cite{ba2016using} & 0.85\% \\
\hline
FW-LN (Dynamic attention) & 0.65\% \\
\hline
WeiNet (Fixated attention) &0.91 \%  \\
\hline
WeiNet (Dynamic attention) &0.49\%  \\
\end{tabular}
\end{table}

\subsection{MNIST Image classification with visual glimpses}

Recurrent Visual attention \cite{mnih2014recurrent}, designed for image classification task, includes a ``glimpses'' network that extracts smaller regions of the image and a recurrent neural network to compose glimpses and generate a compressed representation for classification. 
The visual attention model has to obtain meaningful glimpses from an image, i.e., glimpses that are ``on-target'', and also composes glimpses intelligently, i.e., compose the meaningful glimpses in a way that makes sense for classification. The RNN in the model uses previous glimpses representation to generate a hidden state, both for proposing the next glimpse and for classifying the whole image. Our experiment compares WeiNet with the three RNNs, LSTM, Recurrent Highway Networks and Fast Weights, in Recurrent Visual Attention model by switching the RNN component therein. For WeiNet we use one auto-associative memory and without the router component. All tested RNNs has the same number of hidden units (200) and one recurrent layer for each time step. We evaluated the RNNs within the visual attention framework with two following glimpse settings: 

1). \textit{Dynamic Attention}: We follow \cite{mnih2014recurrent} on using a glimpse network to learn where to extract glimpse in the input image, and use RNN for glimpse composition. 

2) \textit{Fixated Attention}\cite{ba2016using}: We decompose a 28$\times$28 image into four 7$\times$seven fixed glimpse regions without pixel overlap and then compose the glimpses with RNN in the same way as setting 1). 

Most of the implementation details follow \cite{mnih2014recurrent}. But we did not use Reinforcement Learning for training while preserving the glimpse chain sampling. Reasons for this change are: 1) we hypothesis that adding RL reward (based on the final classification error) overlaps with the training criterion (also based on classification error) may not lead to improved accuracy of the model, and 2) we could stress-test the glimpse networks on its capability of generating hidden states and aggregating historical glimpses without the chance for trial and error. Without rewards, the network will have to try hard to avoid generating the wrong chain of glimpses and try to generate a good chain in one shot, which is even harder to learn than RL setting. But we kept Monte Carlo sampling to acquire ten samples (chain of glimpses), and use six glimpses, i.e., 6 RNN steps for each image. We set to use RNN hidden state size to be 200 for all four tested RNNs.

Results of using dynamic attention to compare 4 RNNs are in Table ~\ref{table:va_da}, and comparing fixed and dynamic attention in table \ref{table:va_fa_vs_da}. The results are reported for the test set, and we took six glimpses for dynamic attention setting and four glimpses for fixed attention setting. We use more steps for dynamic attention setting than fixed attention setting is that we would increase the possibility that all informative glimpses to be included. In fixed attention, it is guaranteed to include all pixels, while the hard image-cutting decision may lead to non-informative glimpses where none of the segments may give a hint of what the digit in the image is, as is shown in the Fig. \ref{fig:rva}. It is obvious that fixed attention setting is much harder than dynamic attention setting, which is confirmed in Table 3.

The reason for WeiNet's performance is due to the learned auto-associative memory update since WeiNet has been thinned down to a basic model to be comparable to Fast Weights \cite{ba2016using}. The glimpse network conditions on the RNN hidden state, the only variable therein; The glimpses, in turn, help create a better RNN hidden state for next glimpse proposal. This positive feedback loop between glimpses and RNN states leads to WeiNet's better overall accuracy in Table \ref{table:va_da} than Fast Weights.

We also test WeiNet to use multiple auto-associative memories and the router network, to see if composing parallel memories non-linearly could lead to further improvement. MNIST task is proper for the purpose because glimpses carry different kinds of information: sometimes the glimpse contains a stroke of a digit that is helpful for the classification, while sometimes the glimpse contains nothing useful. Telling apart those useful and useless glimpses should help filter out noisy inputs and lead to better classification performance. And we use two memory cell in the hope to process different kinds of glimpses differently so that the types can be explicitly separated into two dedicated memory cells to be processed. 

We use four loops on a two auto-associative-memory WeiNet as is shown in Figure ~\ref{fig:rva}, and within each inner loop, the fixed glimpse or dynamically generated glimpse representation is fed. In this case, the router component is used to drive glimpses to different memory cells. However, using multiple memory cells creates a model non-identifiability problem, which is why learning such kind of softmax-based memory addressing mechanism is highly unstable, as is observed in Neural Turing Machines (initializing parameters differently may lead to different converged models, sometimes to non-converged model).  To account for non-identifiability, we guide the router to distribute ''informative glimpses'' into one memory, and non-informative glimpses into another. To tell apart which is informative and which is not, we use simple heuristics to calculate how many pixels in the glimpse are non-black pixels. Then we set a threshold for the number of pixels, $p=10$ to determine how much non-black pixels a glimpse should have to be accepted as an informative glimpse. $p$ is determined by cross-validation.

Additionally, we initialized $\mathbf{A}^k$ with two Gaussian $\sigma(0.1, 1)$ and $\sigma(0.9,1)$ to further differentiate the memory content, which may help further distinguish the two memory content. Also, Each glimpse is fed into the router as additional input, and function $g$ in eq. 3 is changed into $g(A_t, \mathbf{w} \odot \mathbf{h}_t + \mathbf{V} \mathbf{g}_t)$ where $g_t\in \mathbf{R}^{G}$ is glimpse vector with size $G$ generated from glimpse networks, and $V \in \mathbf{R}^{G\times H} ,\mathbf{w} \in \mathbf{R}^H$. 
When using dynamic attention with two memory cells, WeiNet could achieve 0.45\% accuracy, which is 8\% improvement on error rate over using single auto-associative memory. When using more than three memory cells, the test accuracy does not increase, which may be explained by the increased network capacity that makes training harder, and the difficulty to find proper thresholds to separate the glimpses into more than two types.
We also tried WeiNet on Machine Comprehension task, which results we report in the appendix.

\section{Conclusion}
This paper proposed a learned auto-associative memory update within RNN that shows promise in encoding long sequences and intelligent input composition. The WeiNet introduces multiple auto-associative memory in the network to encode more complex input patterns. As a working memory architecture.

\section{Acknowledgements}
We thank Gerry Tesauro, Kazi Hassan, Matt Reimer,   Mo Yu, Tim Klinger, Yang Yu for the help to this paper.

\bibliography{aaai18}
\bibliographystyle{aaai18}

\appendix
\newpage

\end{comment}

\twocolumn[\section{Appendix}]
\setcounter{figure}{0}    
\setcounter{table}{0}

\section{Associative Recall task Details}

Each character is represented as one-hot encoding. Length $I$ is 37 in this task (26 characters + 10 digits + 1 question mark). Hidden state size $H$ is set to 50.

In equation 5, The initial value for the $A_0$ is set to 0 matrix. The weight matrices $W_A$ and $W_h$ is set to normal distribution $\mathcal{N}(0.9,0.1)$ and $\mathcal{N}(0.5,0.1)$. This is to use the corresponding scalar parameters to initialize (Ba et al., 2016). For $W_AH$, we initialize it randomly by $\mathcal{N}(0,0.1)$. 

The rest of the network parameters are initialized as $\mathcal{N}(0,0.1)$. We use ADAM optimizer with learning rate 1e-4, batch size 128, gradient clipping (-5,5). The RNN step for each experiment is fixed to 9, 30, 50 respectively.

We also experimented using $H=100$ for all other RNNs but WeiNet for sequence length 50, to see if it helps. We also tried FW-LN using more inner loops. The experiment result does not show any of those experiments could succeed. The result stays pretty much the number reported. 

We suspect that for LSTM and RHN, the learning is already too hard. increasing $H$ to a significant larger, even could help, does not make sense to compare to due to large memory requirement. For FW-LN, we suspect the model does not learn any better when $H>100$, and up to 500, and we do not see the model converge either.

\section{Recurrent Visual Attention for MNIST Details}
For all experiments, we set batch size 32 for training, hidden state size $H$ is 256 for all modules' internal representation, number of glimpses for dynamic attention is set to 6. To be comparable to fixated attention setting, we set the glimpse length and height to be both 7. For fixated attention, the image is evenly split to 4 sub-images of the same, non overlapping 7$\times$7 images for left top, right top, left bottom, right bottom conner. We use ADAM optimizer with learning rate 1e-3, the minimum learning rate is set to 1e-4, meaning when rate decay happens, it won't decay to less than 1e-4. Learning rate decay rate is set to 0.98. We use 10 Monte Carlo samples, 

In Figure 4, the whole 28$\times$28 image is input to the controller first, to get a general idea of the whole picture. Then, each ``inner loop'' will accept a 7 $\times$ 7 sub-image that directs Routing mechanism in eq. 3 and eq. 4. Each 7 $\times$ 7 image is transformed into a hidden state $\mathbf{h}_g = W_g x_{glimpse}$, where $W_g \in \mathcal{R}^{49\times H}$, and then used as additional term injected into eq. 4 into 

\begin{align*}
g(\mathbf{A}_t,\mathbf{h}_t) &= \\ 
&[\mathbf{h}_t'\mathbf{A}^1_t\mathbf{h}_t + \mathbf{h_g}_t'\mathbf{A}^1_t\mathbf{h_g}_t, ...,
\mathbf{h}_t'\mathbf{A}^K_t\mathbf{h}_t + \mathbf{h_g}_t'\mathbf{A}^K_t\mathbf{h_g}_t]
\end{align*}

Thus the router $R$ will be aware of both the full image, as well as the glimpse. 

\section{Machine Reading Comprehension to test WeiNet when Attention is Present}
We show how WeiNet performs compared to other RNNs in a setting where Attention mechanism is present. 
Extractive Machine comprehension is question answering task where a document is provided for extracting answers. Best performing neural networks should be able to properly encode passage and question words with their context information, and then match each passage word representation with the question word representation to form ``attention''. 
\\
\\
We choose Stnaford Question Answering Dataset and a popular model Bi-directional attention flow model for reference architecture. A general structure of such kind of model is illustrated in appendix Figure \ref{fig:rc}. We use the publicly available code from the author as of Jan 18, 2017, And simply replace the RNNs for both passage and questions. The parameters are all default setting. Note that we use batch size 2 for our experiments.
\\
\\
Evidence from our MNIST experiment shows that even when the input sequence is short, WeiNet still could bring additional improvement over other reference RNNs. This is due to the reason that WeiNet could help input representation even in short-input setting. In reading comprehension model when attention is present, the attention mechanism does not explicitly encourage long-input encoding, because words local context are matched more easily with attention mechanism, thus relaxes the LSTM's job from discovering long-term dependencies to local context. Thus we think that even when attention mechanism is present, WeiNet should further improve the performance. The experimental result is shown in appendix table \ref{table:rc_result}.
\\
\\
Note that we do NOT intend to compare with the state of the art on SQuAD leaderboard, and the model structure we applied is NOT yet tuned exhaustively. The performance however, is largely limited by the base model where RNN encoders for passage and question are replaced. 
\begin{figure}
\centering
\begin{minipage}[t]{.45\textwidth}
\centering
\vspace{0pt}
\includegraphics[width=40mm,scale=0.4]{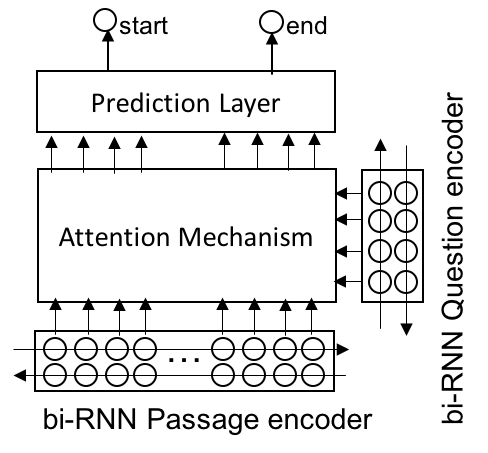}
\caption{A general architecture on Machine Reading Comphrenension}
\label{fig:rc}
\end{minipage}\hfill
\begin{minipage}[t]{.45\textwidth}
\centering
\vspace{0pt}
\captionof{table}{Comparison of WeiNet with other RNN architectures in bi-attention-flow framework for RC}
\label{table:rc_result}
\begin{tabular}{c|c|c}
\hline
  & Dev set EM & Dev set F1 \\
\hline
LSTM  & 65.9\% &  75.7\% \\
\hline
FW-LN  & 64.8\% &  75.3\% \\
\hline
RHN  & 65.3\% &  75.8\% \\
\hline
WeiNet & 67.4\% & 76.8\% \\
\end{tabular}
\end{minipage}
\end{figure}


\end{document}